\documentclass[conference]{IEEEtran}
\IEEEoverridecommandlockouts
\usepackage{cite}
\usepackage{amsmath,amssymb,amsfonts}
\usepackage{algorithmic}
\usepackage{graphicx}
\usepackage{textcomp}
\usepackage{xcolor}
\usepackage{verbatim}

\def\BibTeX{{\rm B\kern-.05em{\sc i\kern-.025em b}\kern-.08em
    T\kern-.1667em\lower.7ex\hbox{E}\kern-.125emX}}
\begin{document}

\title{Probabilistic electrical power demand forecasting with uncertainty quantification\\
}

\author{\IEEEauthorblockN{ Mahesh Neupane}
\IEEEauthorblockA{\textit{Department of Information Technology} \\
\textit{Nepal College of Information  Technology}\\
Lalitpur, Nepal
 \\
mahesh.neupane@ncit.edu.np}
\and
\IEEEauthorblockN{Pragya Dhungana}
\IEEEauthorblockA{\textit{Interconnection and Numbering Section} \\
\textit{Nepal Telecommunications Authority}\\
Kathmandu, Nepal \\
pdhungana@nta.gov.np }    
\and
\IEEEauthorblockN{Pradip Khatri}
\IEEEauthorblockA{\textit{} 
\textit{Nepal Electricity Authority}\\
Kathmandu, Nepal \\
pradipkhatri@pcampus.edu.np}
\and
\IEEEauthorblockN{Swechhya Baskota}
\IEEEauthorblockA{\textit{Department of Global Public Health and Primary Care} \\
\textit{University of Bergen}\\
Bergen, Norway \\
Swechhya.Baskota@student.uib.no }
\and
\IEEEauthorblockN{Hariom Dhungana}
\textit{Western Norway University of Applied Sciences}\\
Bergen, Norway\\
hdhu@hvl.no }   

\maketitle

\begin{abstract}
The majority of research on electricity consumption forecasting has focused on deterministic approaches, which generate a single point estimate for each time step in the forecasting horizon. However, the increasing penetration of renewable energy sources and the growing complexity of modern smart grids have introduced greater variability and uncertainty into power-system demand and operation. Consequently, probabilistic forecasting, which quantifies the uncertainty and variability associated with future electricity demand, is becoming increasingly important for reliable power-system planning and operation. This study presents an empirical comparison of four contemporary probabilistic forecasting models for electricity consumption, highlighting their respective strengths and limitations. 
We have performed comparision on real-world power systems related datasets. Across all power-consumption zones, NGBoost demonstrates superior probabilistic forecasting performance, achieving the lowest MAE and RMSE while providing well-calibrated uncertainty estimates with high prediction-interval coverage and reasonably narrow intervals. These results indicate that NGBoost offers a more accurate and reliable forecasting framework than Bayesian, Monte Carlo (MC) Dropout, and Gaussian Process Regression (GPR) models for the considered electricity consumption data.
\end{abstract}


\section{Introduction}

Accurate forecasting of electricity demand and supply is increasingly important for modern power systems, particularly with the rapid integration of renewable energy sources such as solar and wind \cite{Islam2020EnergyForecasting}. Their weather-dependent variability, combined with dynamic and uncertain consumer behaviour, introduces significant uncertainty into grid operation \cite{Lindberg2023Day-aheadVerification}. Reliable forecasting is therefore essential for effective grid management, resource allocation, system stability, cost reduction, and long-term energy planning.

Forecasting approaches are broadly classified as deterministic and probabilistic \cite{Xie2023AnEnergy}. Deterministic methods provide a single point prediction based on observed data and model assumptions \cite{Miraki2024ElectricityNetwork}, offering simplicity and computational efficiency but providing limited information about forecast uncertainty \cite{Abbasipour2021ATechnique}. In contrast, probabilistic forecasting characterizes multiple possible future outcomes and their associated probabilities \cite{Liu2017ProbabilisticForecasts, Lin2022Short-termMechanism}, thereby supporting risk-aware decision-making. Probabilistic methods can be further categorized as parametric, which assume a predefined probability distribution, and non-parametric, which make fewer distributional assumptions. Uncertainty quantification (UQ) complements these approaches by characterizing the reliability and variability of predictions, including aleatoric and epistemic uncertainty. Although UQ enables more informed planning, risk assessment, and model evaluation, it may introduce additional computational cost and depends strongly on model quality and the accurate characterization of uncertainty sources.

Quantile Regression (QR) is a probabilistic forecasting technique that extends classical regression by estimating conditional quantiles of electricity demand rather than just the mean \cite{Abedinia2024WindApproach}. For a given quantile $\tau \in (0,1)$, QR models the relationship between input features $\mathbf{x}$ and the $\tau$-th conditional quantile $Q_y(\tau \mid \mathbf{x})$ using quantile-specific coefficients, which are estimated by minimizing the asymmetric pinball loss. By predicting multiple quantiles, QR provides a full characterization of the conditional distribution of power consumption, enabling the construction of prediction intervals and explicit uncertainty quantification. This makes it robust to outliers and non-Gaussian noise, and particularly useful in power systems for anticipating demand variability, capturing extreme load events, and supporting risk-aware operational decisions.

Accurate and timely imputation of missing IoT measurements is essential to preserve temporal continuity and statistical consistency of sensor data, thereby enabling more reliable power demand forecasting and reducing bias introduced by incomplete observations \cite{Dhungana2025DataGuidelines}. The accuracy of probabilistic electrical power demand forecasting is influenced by various external factors, such as weather, holidays, and economic data. Traditional methods often incorporate one or more of these factors into models based on expert knowledge, without carefully selecting the most relevant variables. While including more input variables can provide additional reference information, using too many can lead to practical challenges, particularly for neural networks, which are highly sensitive to the scale of input data. Probabilistic electrical demand forecasting enables more transparent and reliable predictions. Specifically::

\begin{enumerate}
    \item \textbf{Uncertainty Quantification:} It captures both epistemic uncertainty (model-related) and aleatory uncertainty (inherent variability in demand), allowing operators to understand forecast confidence. 
    \item \textbf{Risk-Informed Decision Making:} Grid management and planning benefit from probabilistic forecasts by assessing the likelihood of extreme events, such as peak demand or supply shortfalls.
    \item \textbf{Model Evaluation:} Forecast quality can be rigorously evaluated using proper scoring rules like the Continuous Ranked Probability Score (CRPS), facilitating fair comparisons among different models.
    \item \textbf{Support for Bayesian Methods:} Probabilistic forecasts naturally integrate with Bayesian approaches, enabling continuous model updating and improved inference in dynamic power systems.
\end{enumerate}

This work presents a comprehensive, high-level comparison of the various load forecasting techniques discussed, highlighting their methodological differences, strengths, and limitations. The remainder of the paper is organised as follows: Section 2 reviews the concepts behind different probabilistic forecasting methodologies. Section 3 describes the example case, including the dataset and evaluation metrics. Section 4 provides a comparative analysis of the forecasted results, and finally, Section 5 summarises the main findings, outlines future research directions, and concludes the paper.

\section{ Probabilistic forecasting}
Probabilistic forecasting refers to the process of predicting future events by estimating the full probability distribution of possible outcomes and quantifying uncertainty by assigning likelihoods to a range of outcomes. This approach is essential in electrical demand forecasting where systems are inherently stochastic, data are limited, or model uncertainty is significant. Electricity demand in urban power networks is driven by multiple dynamic factors, including temporal patterns (hourly, daily, and seasonal cycles) and exogenous influences such as meteorological conditions. While temporal features alone capture intrinsic consumption rhythms and autocorrelation structures, they cannot fully represent weather-induced variability that directly affects heating and cooling loads. Therefore, this work evaluates demand forecasting using both temporal features and key meteorological variables (temperature, humidity, wind speed, diffuse solar radiation, and global diffuse flows). This design allows assessment of the incremental contribution of weather inputs, where models using only temporal features are expected to show reduced accuracy, particularly during extreme or seasonally variable conditions. 

The input features are normalised using StandardScaler, which transforms each feature to zero mean and unit variance to ensure numerical stability and balanced contribution during model training. This preprocessing step mitigates scale-related bias among heterogeneous features while preserving the relative variability and physical interpretability of the measurements. To prevent data leakage, the scaling parameters were fitted exclusively on the training data. The same fitted scaler was then applied without refitting to the validation and test sets.

A probabilistic forecast can be formally represented as a predictive distribution:
\begin{equation}
    p(y_{t+k} \mid \mathcal{F}_t),
\end{equation}
where $y_{t+k}$ denotes the future value of interest at horizon $k$, and $\mathcal{F}_t$ represents all information available at time $t$. The distribution $p(\cdot)$ expresses not only the expected value but also the spread and tail behaviour of the forecast.

These four models are selected as they span complementary probabilistic frameworks: parametric Bayesian Linear Regression (BLR), approximate Bayesian deep learning (Monte Carlo Dropout), non-parametric kernel-based Gaussian Process Regression (GPR), and distributional boosting named Natural Gradient Boosting (NGBoost). This enables a systematic comparison of uncertainty quantification and predictive performance for power demand forecasting.

\subsection{ Bayesian Linear Regression }

BLR is a probabilistic extension of classical linear regression in which model parameters are treated as random variables rather than fixed but unknown quantities \cite{Fumo2015RegressionConsumption}. Given an input vector $\mathbf{x} \in \mathbb{R}^d$, the target variable is modeled as
\[
y = \mathbf{w}^\top \mathbf{x} + \varepsilon,
\]
where $\mathbf{w}$ denotes the regression coefficients and $\varepsilon \sim \mathcal{N}(0, \sigma^2)$ represents Gaussian noise. In the Bayesian framework, a prior distribution is placed over the parameters, commonly a multivariate Gaussian prior $\mathbf{w} \sim \mathcal{N}(\mathbf{0}, \alpha^{-1}\mathbf{I})$. Using Bayes’ theorem, the posterior distribution of the weights given the data $\mathcal{D} = \{\mathbf{X}, \mathbf{y}\}$ is obtained as
\[
p(\mathbf{w} \mid \mathcal{D}) = \frac{p(\mathbf{y} \mid \mathbf{X}, \mathbf{w}) \, p(\mathbf{w})}{p(\mathbf{y} \mid \mathbf{X})},
\]
which results in a closed-form Gaussian posterior due to conjugacy. This posterior captures parameter uncertainty explicitly, a key advantage over frequentist linear regression.

A major strength of BLR lies in its predictive distribution, which naturally quantifies uncertainty in predictions. For a new input $\mathbf{x}_*$, the predictive distribution is given by
\[
p(y_* \mid \mathbf{x}_*, \mathcal{D}) = \int p(y_* \mid \mathbf{x}_*, \mathbf{w}) \, p(\mathbf{w} \mid \mathcal{D}) \, d\mathbf{w},
\]
resulting in a Gaussian distribution whose mean represents the expected prediction and whose variance reflects both data noise and model uncertainty. This makes BLR particularly suitable for applications where uncertainty estimation is critical like in energy consumption forecasting.

\subsection{ Monte Carlo (MC) Dropout } 
MC Dropout is a practical and widely used probabilistic deep learning approach that enables uncertainty estimation in neural networks without modifying their underlying architecture \cite{Serpell2019ProbabilisticNetworks}. By retaining dropout layers during inference and performing multiple stochastic forward passes, MC Dropout approximates Bayesian inference in deep neural networks. Given an input $\mathbf{x}$, the predictive distribution is estimated by sampling $T$ outputs $\{ \hat{y}_t \}_{t=1}^{T}$ from the network with dropout activated, such that
\[
p(y \mid \mathbf{x}, \mathcal{D}) \approx \frac{1}{T} \sum_{t=1}^{T} p(y \mid \mathbf{x}, \mathbf{W}_t),
\]
where $\mathbf{W}_t$ denotes the randomly dropped network weights at the $t$-th forward pass.

\subsection{ Gaussian Process Regression} 
GPR  is a non-parametric  technique that predicts a distribution over functions consistent with observed data, providing both mean predictions and uncertainty estimates \cite{Williams1995GaussianRegression}. The main idea is to define a Gaussian process prior over the latent function $f(x)$ such that any finite set of function values follows a multivariate normal distribution:

\begin{equation}
f(x) \sim \mathcal{GP}\big(m(x), k(x, x')\big)
\end{equation}
where $m(x)$ is the mean function (often assumed zero) and $k(x, x')$ is the covariance (kernel) function encoding similarity between inputs $x$ and $x'$. Given training data $\{X, y\}$, the predictive distribution for a new input $x_*$ is Gaussian with mean and variance:
\begin{align}
\mu_* &= k(x_*, X) [K(X, X) + \sigma_n^2 I]^{-1} y, \\
\sigma_*^2 &= k(x_*, x_*) - k(x_*, X) [K(X, X) + \sigma_n^2 I]^{-1} k(X, x_*),
\end{align}
where $K(X, X)$ is the covariance matrix of training points, $k(x_*, X)$ is the covariance vector between the test point and training points, and $\sigma_n^2$ is the noise variance.

\subsection{ Natural Gradient Boosting } 
NGBoost extends gradient boosting to estimate full predictive distributions instead of only point predictions \cite{Duan2020NGBoost:Prediction} . The main idea is to model the conditional distribution of the target variable $y$ given features $x$ by parameterising a chosen probability distribution $p(y \mid x; \theta)$ and iteratively learning its parameters $\theta$ using \textit{natural gradients}. At each boosting iteration, NGBoost fits a base learner, typically a decision tree, to the natural gradient of a proper scoring rule, such as the negative log-likelihood, with respect to the distribution parameters.

Mathematically, for the parameter $\theta$ of the predictive distribution, the update rule is:
\begin{equation}
\theta_{t+1} = \theta_t - \eta \, F^{-1} \nabla_\theta L(y, p(y\mid x; \theta_t))
\end{equation}
where $L$ is the negative log-likelihood loss, $\nabla_\theta L$ is the gradient, $F^{-1}$ is the inverse Fisher information matrix defining the natural gradient, and $\eta$ is the learning rate.

\section{Example cases}

\subsection{Dataset}\label{AA}
The electricity demand dataset used in this study was collected from three major power distribution networks in Tetouan city, located in northern Morocco \cite{Salam2018ComparisonCity}. The data were sourced from the Supervisory Control and Data Acquisition (SCADA) system operated by Amendis, the public utility responsible for electricity and water distribution in the region. The distribution network comprises three primary supply stations: Zone~1 (Quads), Zone~2 (Smir), and Zone~3 (Boussafou). The original dataset contains 52,416 observations across the three zones, recorded at 10-minute intervals over the full year of 2017. Prior to resampling, the dataset was checked for data-quality issues. No missing values, duplicate records, outliers, or anomalous readings were identified; therefore, no imputation, removal, or correction was required. The data were subsequently resampled to hourly intervals by aggregating the 10-minute measurements, thereby reducing short-term fluctuations while preserving the overall consumption patterns.
Table~\ref{tab:data} summarises the input variables and target outputs, including temporal, meteorological, and flow-related features used for power consumption forecasting.

\begin{table}[htbp]
\caption{Variables used for forecasting power consumption.}
\label{tab:data}
\centering
\begin{tabular}{|c|c|c|}
\hline
\textbf{Variable} & \textbf{Variable Description} & \textbf{Units} \\
\hline
Datetime & Hour & min \\
         & Day & -- \\
         & Month & -- \\
\hline
Temperature & Weather temperature & $^\circ\mathrm{C}$ \\
Humidity & Weather humidity & \% \\
WindSpeed & Wind speed & km/h \\
GeneralDiffuseFlows & General diffuse flow & kg/s or L/s \\
DiffuseFlows & Diffuse flow & kg/s or L/s \\
\hline
PowerConsumption\_Zone1 & Power demand & kW \\
PowerConsumption\_Zone2 & Power demand & kW \\
PowerConsumption\_Zone3 & Power demand & kW \\
\hline
\end{tabular}
\end{table}

\begin{figure}[htbp]
\centering
\includegraphics[width=\columnwidth]{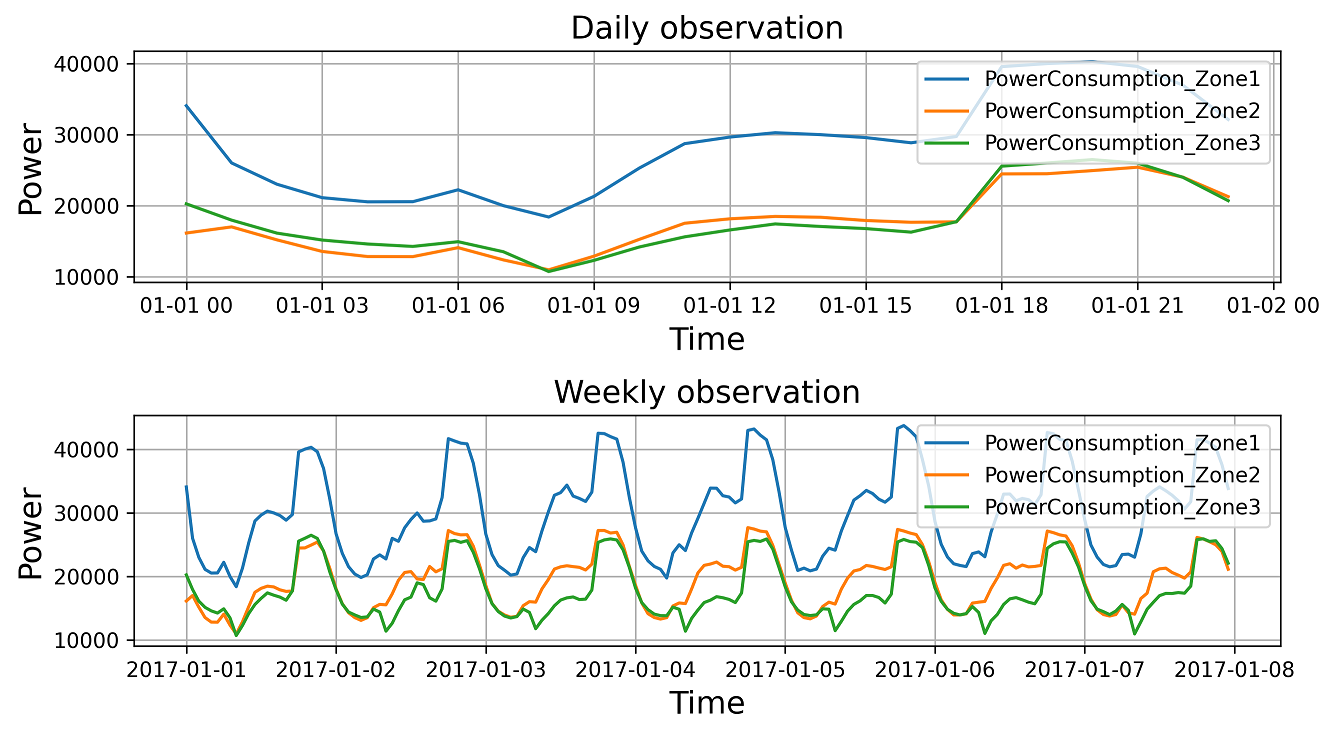}
\caption{Electricity demand is visualised for the distribution scale. }  \label{fig:pattern}
\end{figure}

Figure~\ref{fig:pattern} illustrates daily and weekly power consumption patterns across three zones (Zone~1, Zone~2, and Zone~3). The daily profile (top panel) shows that Zone~1 consistently records the highest demand, decreasing to a minimum around 09:00 before rising to a peak in the late afternoon and evening. Zones~2 and~3 exhibit similar temporal trends but at considerably lower demand levels, closely following each other with comparable minima and peak periods. The weekly profile (bottom panel), covering the first week of January 2017, reveals a pronounced daily cyclic behaviour across all zones, reflecting typical residential and commercial activity patterns. Zone~1 continues to dominate overall consumption with greater variability, while Zones~2 and~3 maintain lower and more stable demand throughout the week.

\begin{figure}[htbp]
\centering
\includegraphics[width=\columnwidth]{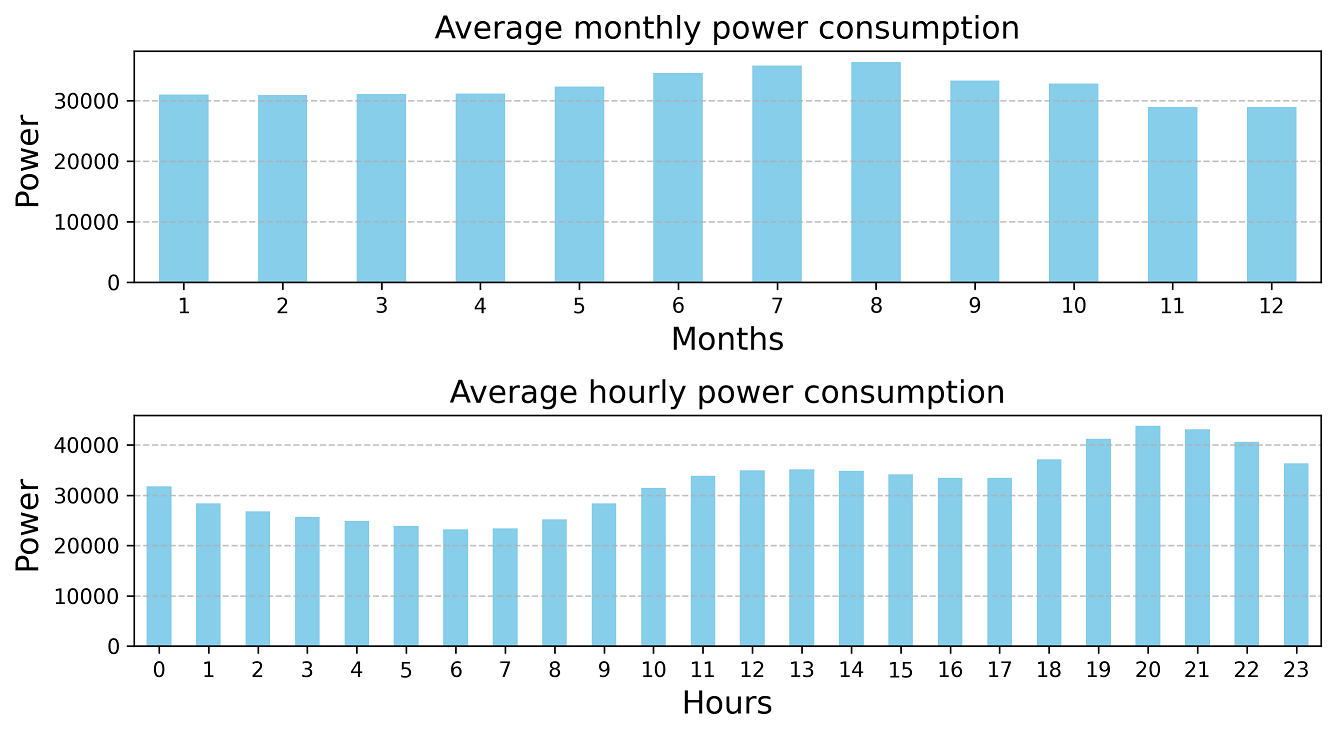}
\caption{Average power consumption pattern across various month and different hours} \label{fig:average}
\end{figure}

Figure~\ref{fig:average} shows average power consumption across monthly and hourly time scales.The top panel presents average monthly consumption, which remains relatively stable throughout the year at approximately $30{,}000$ units, with a modest increase during the summer months (June–August). The bottom panel illustrates average hourly consumption over a 24-hour period, revealing lower demand during early morning hours (02:00–07:00) and a steady rise toward a pronounced evening peak around 20:00–21:00.

\subsection{Evaluation metric}

To comprehensively assess the performance of probabilistic regression models, seven evaluation metrics are employed, capturing both predictive accuracy and uncertainty quality. They can be evaluated in terms of consistency, variation, and the tightness of the estimated distribution.  Mean Absolute Error (MAE) measures the average magnitude of prediction errors by computing the mean of the absolute differences between the true values and the predicted values, without penalising large errors disproportionately \cite{Dhungana2025APlanning}. RMSE evaluates point prediction accuracy using the predictive mean, and Negative Log-Likelihood (NLL) measures how well the predicted probability distribution fits the observed data; lower is better \cite{Salzmann2022Motron:Forecasting}. The CRPS evaluates the accuracy of probabilistic forecasts by measuring the difference between the predicted cumulative distribution and the observed outcome, capturing both reliability and sharpness. The pinball score assesses model calibration, reliability, and sharpness by evaluating quantile prediction accuracy. It is calculated as the mean of pinball losses over the entire forecast horizon and across all specified quantiles. Prediction Interval Coverage Probability (PICP) measures the proportion of true values contained within the predicted intervals, indicating interval reliability, with values close to the nominal confidence level reflecting better uncertainty calibration and sharpness \cite{Liu2021ProbabilisticNetwork}. The Mean Prediction Interval Width (MPIW) measures the sharpness of predictive uncertainty; the narrower it is, the better.
The formulas of these metrics are shown in  \eqref{eq:mae} - \eqref{eq:mpiw}. 

\begin{equation}
\mathrm{MAE} = \frac{1}{N}\sum_{i=1}^{N} \left| y_i - \mu_i  \right|    \label{eq:mae}
\end{equation}

\begin{equation}
\mathrm{RMSE} =\sqrt{ \frac{1}{N} \sum_{i=1}^{N} \left( y_i - \mu_i \right)^2 } \label{eq:rmse}
\end{equation}

\begin{equation}
\mathrm{NLL} = \frac{1}{N} \sum_{i=1}^{N} \left[ \frac{(y_i - \mu_i)^2}{2\sigma_i^2}
+ \frac{1}{2} \log \left( 2\pi\sigma_i^2 \right) \right]
\end{equation}

\begin{equation}
\mathrm{CRPS}(\mu_i, \sigma_i, y_i) = \sigma_i \left[ z_i \left( 2\Phi(z_i) - 1 \right)
+ 2\phi(z_i) - \frac{1}{\sqrt{\pi}}  \right] \quad  \end{equation}
\[
\text{Where: } z_i = \frac{y_i - \mu_i}{\sigma_i}.
\]

\begin{equation}
\mathcal{L}_{\tau}(y_i, \hat{y}_i) = \begin{cases} \tau (y_i - \hat{y}_i), & y_i \ge \hat{y}_i, \\
(1 - \tau)(\hat{y}_i - y_i), & y_i < \hat{y}_i.  \end{cases}
\end{equation}

\begin{equation}
\mathrm{PICP} = \frac{1}{N} \sum_{i=1}^{N} \mathbb{I} \left( L_i \le y_i \le U_i \right).
\end{equation}

\begin{equation}
\mathrm{MPIW} = \frac{1}{N} \sum_{i=1}^{N} \left( U_i - L_i \right). \label{eq:mpiw}
\end{equation}

Where: $U_i$ and  $L_i$  are Upper and  Lower bounds are obtained by subtracting and adding the scaled predictive standard deviation to the mean, capturing the range in which the true value is expected to lie with probability.  A Gaussian prediction interval with confidence level $1-\alpha$ is constructed to denote the significance level. Here, we use $\alpha = 0.05$, which corresponds to a $95\%$ prediction interval.

\section{Results and discussions}
This section describes the experimental setup, including the data partitioning strategy for training and testing. The original data were resampled to an hourly resolution to ensure temporal consistency. The dataset was then split chronologically, with the first 80\% used for training and the remaining 20\% reserved for testing. MC Dropout is realized by enabling dropout layers during prediction and aggregating multiple outputs to compute the predictive mean and variance in a TensorFlow-based notebook implementation. The predictive mean represents the final model estimate, while the variance provides a quantitative measure of epistemic uncertainty.

\begin{figure}[htbp]
\centering
\includegraphics[width=\columnwidth]{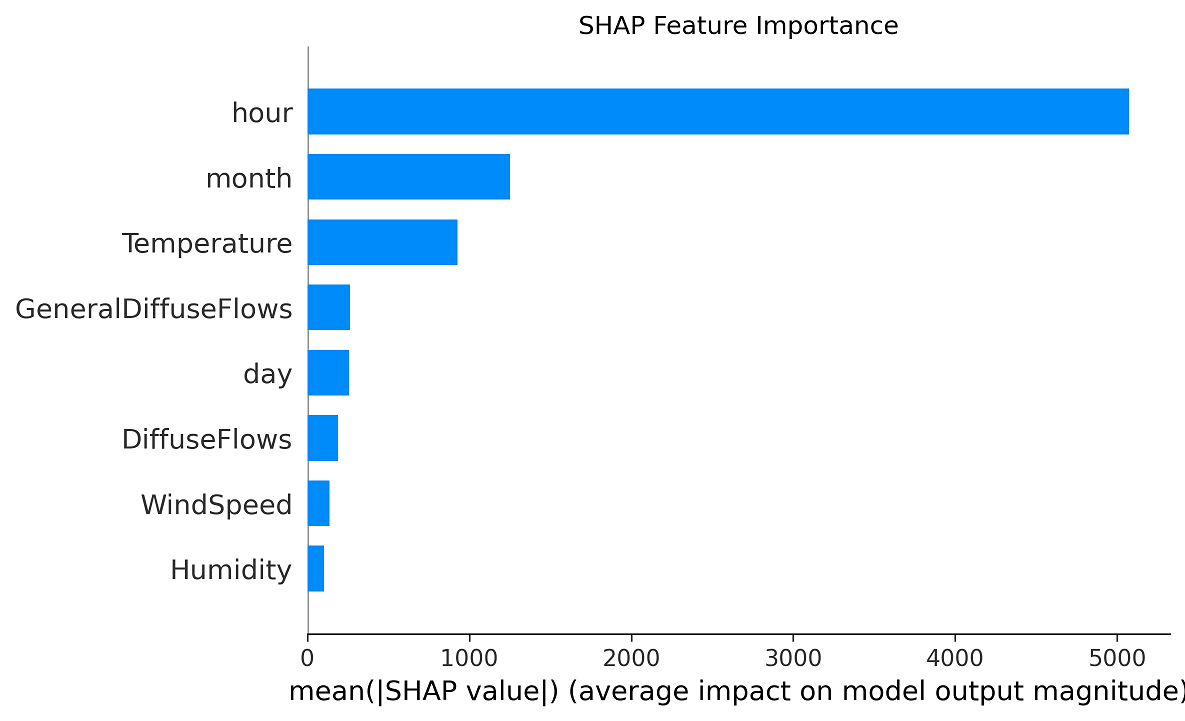}
\caption{SHAP feature importance for power consumption prediction using the XGBoost model} \label{fig:shap}
\end{figure}

Figure~\ref{fig:shap} displays the mean absolute SHAP feature importance value (average impact on model output magnitude) for various features used in a power consumption prediction model. The 'hour' feature is overwhelmingly the most important predictor, with a mean SHAP value exceeding 5000, suggesting that the time of day has the largest impact on power consumption magnitude. The 'month' feature is the second most important, followed by 'Temperature'. Features related to weather and calendar, such as 'GeneralDiffuseFlows', 'day', 'DiffuseFlows', 'WindSpeed', and 'Humidity', have significantly lower but still non-zero importance, with their mean SHAP values all below 1000, demonstrating that while they contribute to the prediction, their average impact is far less than that of the time and temperature variables.

The results in Table~\ref{tab:result} demonstrate a clear distinction in the performance of probabilistic prediction models across different power consumption zones. Overall, NGBoost consistently achieves the lowest deterministic errors (MAE and RMSE) and strong probabilistic scores (low NLL, CRPS, and Pinball Loss), while maintaining high interval coverage  $(\mathrm{PICP} \approx 0.95\text{--}0.96)$  with reasonably narrow prediction intervals (MPIW). Bayesian and MC Dropout models show moderate predictive accuracy and reliable uncertainty estimates, with MC Dropout slightly outperforming Bayesian methods in MAE and RMSE across most zones. 

GPR demonstrates substantially higher errors and near-zero coverage $(\mathrm{PICP} \approx 0)$, indicating both inaccurate mean predictions and severely miscalibrated uncertainty estimates. This poor performance can be attributed to the high dimensionality and non-stationary characteristics of power-demand data, which challenge kernel-based methods that rely on stationarity assumptions and are sensitive to kernel and hyperparameter selection. Moreover, the cubic computational complexity of GPR limits its ability to scale effectively with large datasets, often resulting in over-smoothed predictions and overly narrow predictive intervals that fail to capture true demand variability. Consequently, GPR underperforms relative to NGBoost, which better handles complex nonlinear relationships and heteroscedastic uncertainty in large-scale energy forecasting tasks.

Across all zones, the Bayesian model exhibits the lowest training time $( \approx 0.01-0.05~s )$, which is expected due to its closed-form parameter updates and convex optimization, leading to rapid convergence. In contrast, MC Dropout and NGBoost require iterative stochastic optimization and ensemble boosting, respectively, resulting in moderate training times $( \approx 22-63 s)$ that reflect repeated forward--backward passes and sequential learner fitting. The substantially higher training time of GPR $( \approx 75-89 s)$ is scientifically justified by its cubic computational complexity $\mathcal{O}(n^3)$ associated with kernel matrix inversion, which becomes increasingly expensive with larger training samples.

\begin{table*}[htbp]
\caption{ Comparative performance of probabilistic prediction models across different power consumption zones, evaluated using deterministic accuracy metrics (MAE, RMSE) and proper probabilistic scoring and interval quality measures (NLL, CRPS, Pinball Loss, PICP, and MPIW) }
\begin{center}
\begin{tabular}{|c|c|c|c|c|c|c|c|c|c|}
\hline
\textbf{Zone} & \textbf{Technique} & \textbf{Traine time (sec)} & \textbf{\textit{MAE}}& \textbf{\textit{RMSE}}& \textbf{\textit{NLL }}& \textbf{\textit{CRPS}}  & \textbf{\textit{Pinball}}& \textbf{\textit{PICP }}& \textbf{\textit{MPIW}}\\
\hline
& Bayesian  &  0.0284 & 3454.93 & 4306.62 & 9.79 &  2423.37 &  1727.46  & 0.96  & 16792.41 \\ 		
Zone 1 & MC Dropout  & 57.5283 & 3014.88 & 3837.81 &  9.69 &  2137.33  &  1510.81  & 0.92  &  12590.50 \\ 
& GPR & 76.4913 &   32366.05 & 33117.37  & 5490.48  & 32187.64  & 16183.03  & 0.00  & 1239.59  \\
& NGBoost & 22.6066 &  1682.57 &  2241.14  & 9.00  & 1178.65 &  841.29  &  0.95 & 8068.97  \\

\hline
& Bayesian  & 0.0523 & 2659.66 & 3332.76 & 9.53 &  1877.08 &  1329.83  & 0.94  & 13274.41 \\ 		
Zone 2 & MC Dropout  &  63.1979 &  2232.26 &  2843.16 &  9.49 &  1613.57 & 1116.13 &  0.86 &  8100.93  \\ 
& GPR & 89.2137&   21052.16 &  21659.48 &  2352.34  &  20873.74  &  10526.08 &  0.00  &  1239.59 \\
& NGBoost & 23.2634 &   1400.45  &  1799.98  &   8.84 &  981.96 &  700.22 &   0.96  &  6949.29  \\

\hline
& Bayesian  & 0.0109 &3410.88 &  4288.81 &  9.78  &  2404.12 &  1705.44 &  0.95  & 16561.43 \\ 		
Zone 3 & MC Dropout  & 40.3053 & 2496.89 & 3365.98 &  9.88  &  1870.11  &  1248.44 &  0.77  &  7470.49 \\ 
& GPR &  75.9048 & 17949.81 &  19141.04 &  1838.57  &  17771.40  &  8974.90  &  0.00  &  1239.59 \\
& NGBoost &  23.1263 &  1265.71 &  1760.22 &  8.70  &   898.78   &  632.86   &  0.96  &  6601.56   \\

\hline
\end{tabular}   
\label{tab:result}
\end{center}
\end{table*}

\begin{figure}[htbp]
\centering
\includegraphics[width=\columnwidth]{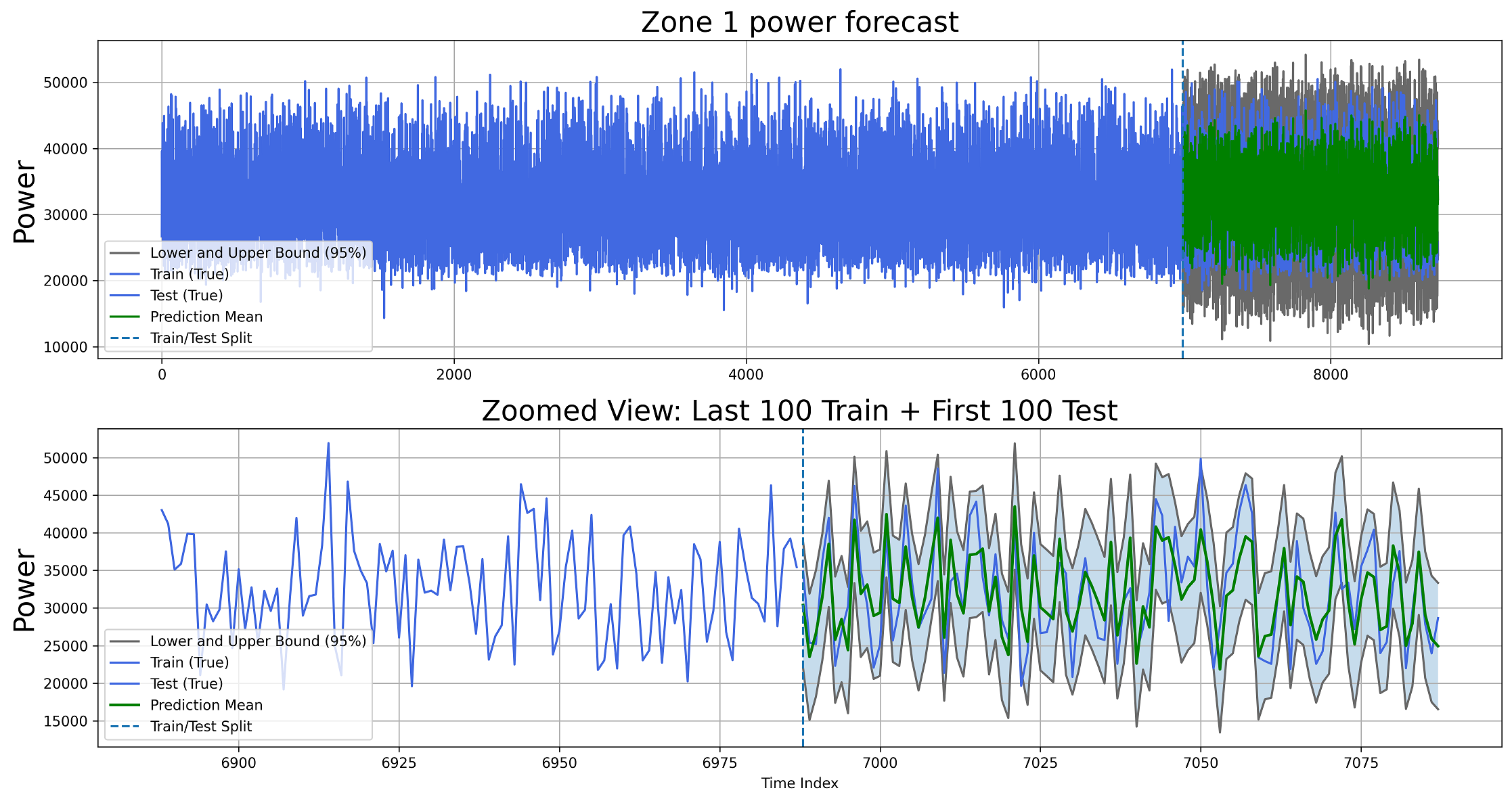}
\caption{Probabilistic Power Consumption Forecasting with NGBoost} \label{fig:forecast}
\end{figure}

Figure~\ref{fig:forecast} presents the results of a probabilistic power consumption forecast model utilizing NGBoost for prediction. The top panel displays the entire time series, with the true consumption data segmented into training (blue) and testing (orange) sets, separated by a dashed blue line. The bottom panel presents a Zoomed View of the Train/Test Split, focusing on the last 100 training values and the first 100 test values to clearly illustrate the model's probabilistic output. In the test region, the Prediction Mean (green line) shows the point forecast. At the same time, the 95\% confidence interval (grey-shaded area, defined by the Upper and Lower Bounds) provides a probabilistic forecast of the range of power consumption. The zoomed view demonstrates that the model successfully captures the cyclical pattern in the consumption data, with the true test values generally falling within the narrow 95\% prediction interval, suggesting the NGBoost model provides the lowest uncertainty estimates for the power consumption forecast.

\begin{figure*}[htbp]
\centering
\includegraphics[width=\textwidth]{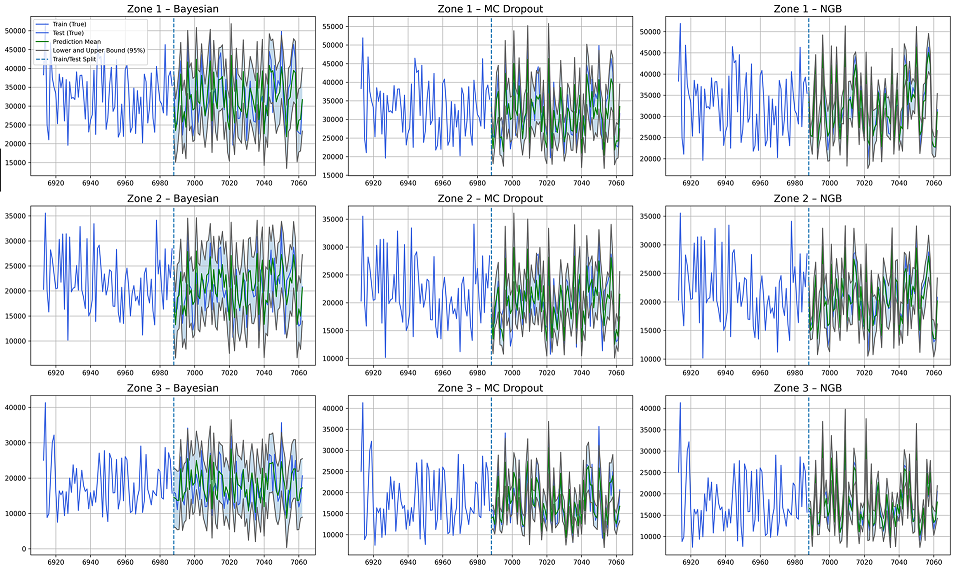}
\caption{Uncertanity quantification plot of three probabilistic forecast models Bayesian, MC Dropout,and NGB Models in three zones.} \label{fig:all}
\end{figure*}

Figure~\ref{fig:all}  presents a 3x3 grid of time series plots of three Zones from Bayesian, MC Dropout, and NGB, respectively. Each plot provides a zoomed-in view of the probabilistic forecasting performance, focusing on the boundary between the training data (blue line) and the testing data (black line), which is demarcated by a vertical dashed blue line. The purpose is to compare the performance of three probabilistic models, BLR, MC Dropout, and NGB, in forecasting power consumption for three different Zones (1, 2, and 3). For the test region, each model displays the Prediction Mean (green line) as the point forecast and the 95\% confidence interval (shaded grey area bounded by the Lower and Upper Bounds) as the measure of predictive uncertainty. The visual analysis of the plots reveals differences in point prediction accuracy and uncertainty quantification across the models and zones. Across all three zones, the NGB model appears to provide the tightest and most responsive confidence intervals, closely tracking fluctuations in the true test data, suggesting that its uncertainty estimates are well calibrated and less conservative than those of the other two. The Bayesian model and MC Dropout model exhibit generally wider confidence bands, particularly for Zone 1, where the power consumption is highest and most volatile, indicating higher estimated uncertainty. For the lower-consumption zones, Zone 2 and Zone 3, all models generally capture the true values within their 95\% intervals, though the NGBoost model consistently maintains a smaller uncertainty band, highlighting its potential superiority in providing sharp and accurate probabilistic forecasts for this energy consumption dataset.

Uncertainty estimates remain highly stable for models with convex training objectives, as their predictive variance is largely insensitive to random initialisation. In contrast, stochastic models exhibit moderate seed-dependent variability due to randomness in weight initialisation, data shuffling, and Monte Carlo sampling.

Across all three zones, the 95\% prediction intervals (grey-shaded bounds) expand during high-demand periods and contract during low-demand periods, indicating a positive correlation between model uncertainty and load magnitude. This pattern reflects heteroscedasticity, where prediction errors increase at extreme demand levels. Model-specific behaviour is also observed: while BLR produces smooth, persistent uncertainty bands, NGBoost exhibits sharper, more localised fluctuations in interval width, highlighting differences in how each model scales variance with demand.

\begin{figure*}[htbp]
\centering
\includegraphics[width=\textwidth]{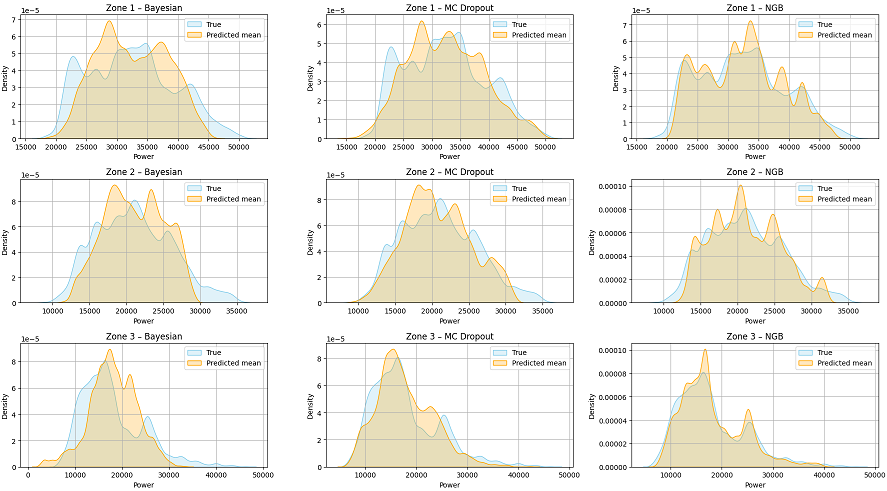}
\caption{Comparison of power consumption forecasting from Bayesian, MC Dropout and NGB based probabilistic Models} \label{fig:kde}
\end{figure*}

Figure~\ref{fig:kde} presents Kernel Density Estimation (KDE) plots of all three Zones from Bayesian, MC Dropout,and NGB, respectively. Each row corresponds to a different consumption Zone (1, 2, or 3), and each column represents a different probabilistic forecasting model. Within each plot, the distribution of the true power consumption values (light blue shaded area) is compared against the distribution of the predicted mean values (orange shaded area). The purpose is to visually assess how well each model captures the overall statistical properties and density shape of the actual power consumption data for each zone. The visual comparison of the KDE plots across the three zones and three models reveals key insights into the models' ability to replicate the underlying data distribution. For Zone 1, all three models show a reasonable alignment between the true and predicted mean distributions, successfully capturing the bimodal nature of the data with peaks around 20,000 and 35,000 units, although the NGBoost prediction appears slightly shifted to the right (higher values). For Zone 2 and Zone 3, the models continue to show good performance, with the NGBoost model generally offering the lowest mean prediction interval width with the true distribution's shape and mode locations compared to the Bayesian and MC Dropout approaches, which sometimes exhibit slightly broader or more skewed predicted distributions, particularly in the lower-consumption ranges. Overall, the NGBoost model appears to be the most effective in replicating the true probability distribution of power consumption across all three zones.

Based on the visual evidence from the KDE plots and the direct forecast comparison, the NGB model consistently provides the most effective probabilistic forecasts for power consumption, demonstrating tighter alignment with the true data distribution and offering the sharpest, best-calibrated 95\% confidence intervals across all three zones compared to the Bayesian and MC Dropout approaches.

The results indicate that BLR provides robust, interpretable probabilistic predictions by incorporating prior knowledge, making it suitable for limited-data scenarios. MC Dropout enhances risk-aware decision-making by providing practical uncertainty estimates, particularly in energy forecasting. NGBoost consistently delivers the most reliable performance, producing well-calibrated probabilistic forecasts with accurate means and sharp prediction intervals while efficiently capturing non-linear relationships. In contrast, GPR effectively models complex non-linearities with principled uncertainty quantification but is constrained by high computational cost for large-scale datasets.

\section{Conclusion and future works}
\paragraph{Conclusion} This study highlights the necessity of probabilistic approaches for forecasting electrical power demand in modern energy systems, where uncertainty and variability play critical roles in reliable planning and operation. By systematically reviewing and empirically comparing state-of-the-art probabilistic forecasting models, the paper demonstrates that probabilistic methods provide substantially richer and more informative predictions than traditional deterministic approaches. The comparative analysis across multiple power consumption zones shows that NGBoost consistently outperforms Bayesian, MC Dropout, and GPR models, achieving the lowest quantitative errors while delivering well-calibrated uncertainty estimates with high prediction interval coverage and sharpness. These results confirm NGBoost as a robust and practical framework for accurate and reliable probabilistic power demand forecasting. Overall, the findings underscore the importance of uncertainty-aware forecasting and offer guidance on selecting probabilistic models for real-world smart grid and energy management applications.

\paragraph{Future works} Future research will extend this study through a comprehensive evaluation of advanced probabilistic forecasting models, including heteroscedastic neural networks and Bayesian deep learning approaches. These methods will be assessed within a unified evaluation framework using proper scoring rules such as CRPS and NLL, complemented by reliability and sharpness metrics. Particular emphasis will be placed on uncertainty calibration and model robustness under diverse operating and weather conditions. Furthermore, the trade-offs among predictive accuracy, uncertainty quality, and computational efficiency will be systematically investigated to support practical deployment in real-world power systems.


\bibliographystyle{ieeetr} %
\bibliography{references}

\end{document}